\documentclass[final]{fcs}

\usepackage{amsfonts}
\usepackage{amssymb}
\usepackage{pifont}
\usepackage{tabularx}
\usepackage{arydshln}
\usepackage{soul}
\usepackage{xcolor}
\usepackage{mathtools}
\usepackage{courier}
\usepackage{adjustbox}
\usepackage{multirow}
\usepackage{CJKutf8}
\usepackage{booktabs}
\usepackage{setspace}
\usepackage{enumerate}
\usepackage{bbding} 
\usepackage{wasysym}
\usepackage{caption}
\usepackage{subcaption}
\usepackage{bbold}
\usepackage{tikz}
\usepackage{graphicx}
\usepackage{enumitem}
\usepackage{url}
\usepackage{hyperref}

\usepackage{times}
\usepackage{latexsym}
\usepackage{tikz}
\usepackage[edges]{forest}
\usepackage{pgfplots}
\usepackage{custom}
\usepackage{amsmath}
\usepackage{bm}
\usepackage{scalefnt}
\usepackage[most]{tcolorbox}

\usepackage[edges]{forest}
\usepackage[framemethod=tikz]{mdframed}
\usepackage{subcaption}

\usepackage[T1]{fontenc}
\usepackage[utf8]{inputenc}
\usepackage{microtype}

\usepackage{inconsolata}

\definecolor{newblue}{RGB}{215,238,249}
\definecolor{neworange}{RGB}{255,243,221}
\definecolor{hidden-black}{RGB}{20,68,106}
\definecolor{tkcolor}{RGB}{255,243,221}
\newtcolorbox{takeaways}[2][]{
	width=\columnwidth,
	colback = tkcolor, 
	colframe = tkcolor, 
	boxsep=0pt,left=10pt,right=10pt,top=0pt,bottom=0pt,
	fontupper=\linespread{0.9}\selectfont,
	title=#2,#1}
\newcommand{\eg}{\textit{e.g.,}}

\usepackage[square,numbers]{natbib}
\title{Large Language Models Meet NLP: A Survey}

\author[1]{Libo Qin}
\author[2]{Qiguang Chen}
\author[3]{Xiachong Feng}
\author[2]{Yang Wu}
\author[1]{Yongheng Zhang} 
\author[4]{\\Yinghui Li}
\author[1]{Min Li}
\author*[2]{Wanxiang Che}
\author[5]{Philip S. Yu}

\address[1]{School of Computer Science and Engineering, Central South University, Changsha 410083, China}
\address[2]{Research Center for Social Computing and Interactive Robotics, \\ Harbin Institute of Technology, Harbin 150001, China}
\address[3]{Department of Computer Science, University of Hong Kong, Hong Kong 999077, China}
\address[4]{Shenzhen International Graduate School, Tsinghua University, Beijing 100084, China}
\address[5]{Department of Computer Science, University of Illinons at Chicago, Chicago 60637‌‌, America}

\fcssetup{
  received       = {month dd, yyyy},
  accepted       = {month dd, yyyy},
  corr-email     = {car@ir.hit.edu.cn},
}
\begin{abstract}
While large language models (LLMs) like ChatGPT have shown impressive capabilities in Natural Language Processing (NLP) tasks, a systematic investigation of their potential in this field remains largely unexplored. This study aims to address this gap by exploring the following questions:
(1) \textit{How are LLMs currently applied to NLP tasks in the literature}? (2) \textit{Have traditional NLP tasks already been solved with LLMs}?
(3) \textit{What is the
future of the LLMs for NLP}?
To answer these questions, we take the first step to provide a comprehensive overview of LLMs in NLP.
Specifically, we first introduce a unified taxonomy including (1) \textit{parameter-frozen paradigm} and (2) \textit{parameter-tuning paradigm} to offer a unified perspective for understanding the current progress of LLMs in NLP. 
Furthermore, we summarize the new frontiers and the corresponding challenges, aiming to inspire further groundbreaking advancements. We hope this work offers valuable insights into {the potential and limitations} of LLMs, while also serving as a practical guide for building effective LLMs in NLP.
\end{abstract}

\begin{document}

\section{Introduction}

Recently, large language models (LLMs) represent a significant breakthrough in AI through scaling up language models~\cite{zhao2023survey,kaddour2023challenges,yang2023harnessing,hadi2023large,zhuang2023through,team2024gemini,guo2025deepseek,chen2024internvl,chen2025ai4research}.
Current studies on LLMs,
such as GPT-series~\cite{brown2020language,ouyang2022training}, PaLM-series~\cite{chowdhery2022palm}, OPT~\cite{zhang2022opt}, and LLaMA~\cite{touvron2023llama}, have shown impressive zero-shot performance. 
In addition, LLMs also bring some emergent abilities including instruction following~\cite{wei2022finetuned}, chain-of-thought reasoning~\cite{wei2022chain}
and in-context learning~\cite{min2022rethinking}, which attract increasing attention~\cite{wei2022emergent}.

\begin{figure}[h]
	\centering
	\includegraphics[width=0.5\textwidth]{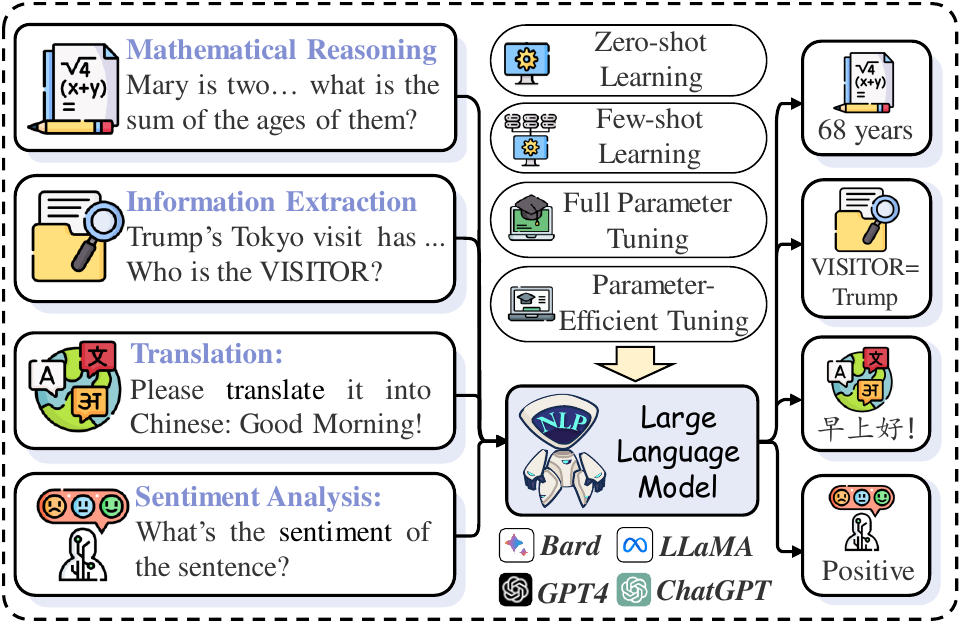}
	\caption{ The example of applying LLMs for NLP tasks (e.g., mathematical reasoning, machine translation, information extraction and sentiment analysis).
	}
	\label{fig:intro}
\end{figure}

With the advancement of large language models, as shown in Figure~\ref{fig:intro}, LLMs allow various natural language processing (NLP) tasks (e.g., zero-shot mathematical reasoning~\citep{wei2022chain,chen2024unlocking}, text summarization~\citep{wang2023zeroshot,wang2023element}, machine translation~\citep{wang2023document,peng2023towards}, information extraction~\citep{wei2023zero,wan2023gpt} and sentiment analysis~\citep{huang_emotionally_2023,wang2023chatgpt}) to be achieved through a unified generative paradigm, which has achieved remarkable success~\cite{zhao2023survey,chen2025towards,zhang2024autocap,ren2025dylas}.
{Additionally, some LLMs in NLP work without needing any additional training data and can even surpass traditional models fine-tuned with supervised learning. This advancement significantly contributes to the development of NLP.}
As a result, the community has witnessed an exponential growth of LLMs for NLP studies, which motivates us to investigate the following questions:
(1) \textit{How are LLMs currently applied to NLP tasks in the literature}? (2) \textit{Have traditional NLP tasks already been solved with LLMs}?
(3) \textit{What is the
future of the LLMs for NLP}?

\newpage

To answer the above questions, we present a comprehensive and detailed analysis on LLMs from the perspective of independent NLP tasks. The overarching goal of this work is to explore current developments in LLMs for NLP.
To this end, in this paper, we first introduce the relevant background and preliminary. Furthermore, we introduce a unified paradigm on LLMs for NLP: (1) \textit{{parameter-frozen paradigm}} including (i) \textit{zero-shot learning} and (ii) \textit{few-shot learning}; (2) \textit{{parameter-tuning paradigm}} containing (i) \textit{full-parameter tuning} and (ii) \textit{parameter-efficient tuning}, {aiming} to provide a unified perspective to understand the current progress of LLMs for NLP tasks:

\begin{itemize}
	\setlength{\itemsep}{1pt}
	\setlength{\parsep}{0pt}
	\setlength{\parskip}{0pt}
	\item \textbf{\textit{Parameter-frozen paradigm}} directly applies prompting approach on LLM for NLP tasks  without the need for parameter tuning.
	This category includes \textit{zero-shot} and \textit{few-shot learning}, depending on whether the few-shot demonstrations is required.

	\item \textbf{\textit{Parameter-tuning paradigm}} refers to the need for tuning parameters of LLMs for NLP tasks. This category includes both \textit{full-parameter} and \textit{parameter-efficient tuning}, depending on whether fine-tuning is required for all model parameters.

\end{itemize}
Finally, we conclude by identifying potential frontier areas for future research, along with the associated challenges to stimulate further exploration.
In summary, this work offers the following contributions:
\begin{itemize}
	\setlength{\itemsep}{3pt}
	\setlength{\parsep}{0pt}
	\setlength{\parskip}{0pt}
	\item[(1)] \textbf{\textit{First survey}}:
	We present the first comprehensive survey of Large Language Models (LLMs) for Natural Language Processing (NLP) tasks.
	\item[(2)] \textbf{\textit{New taxonomy}}: We introduce a new taxonomy including (1) \textit{parameter-frozen paradigm} and (2) \textit{parameter-tuning paradigm}, which provides a unified view to understand LLMs for NLP tasks.
	\item[(3)] \textbf{\textit{New frontiers}}: We discuss emerging areas of research in LLMs for NLP and highlight the challenges associated with them, aiming to inspire future breakthroughs.
	\item[(4)] \textbf{\textit{Abundant resources}}:
	We create the first curated collection of LLM resources for NLP, including open-source implementations, relevant corpora, and a list of research papers. These resources are available at \url{https://github.com/LightChen233/Awesome-LLM-for-NLP}.
\end{itemize}

We hope this work will be a valuable resource for researchers and spur further advancements in the field of LLM-based NLP.

\section{Background}\label{sec:background}
As shown in Figure~\ref{fig:formula}, this section describes the background of 
parameter-frozen paradigm ($\S \ref{sec:parameter-frozen-paradigm}$) and parameter-tuning paradigm ($\S \ref{sec:parameter-tuning-paradigm}$).

\subsection{Parameter-Frozen Paradigm}\label{sec:parameter-frozen-paradigm}
\begin{figure*}[t]
	\centering
	\includegraphics[width=0.95\textwidth]{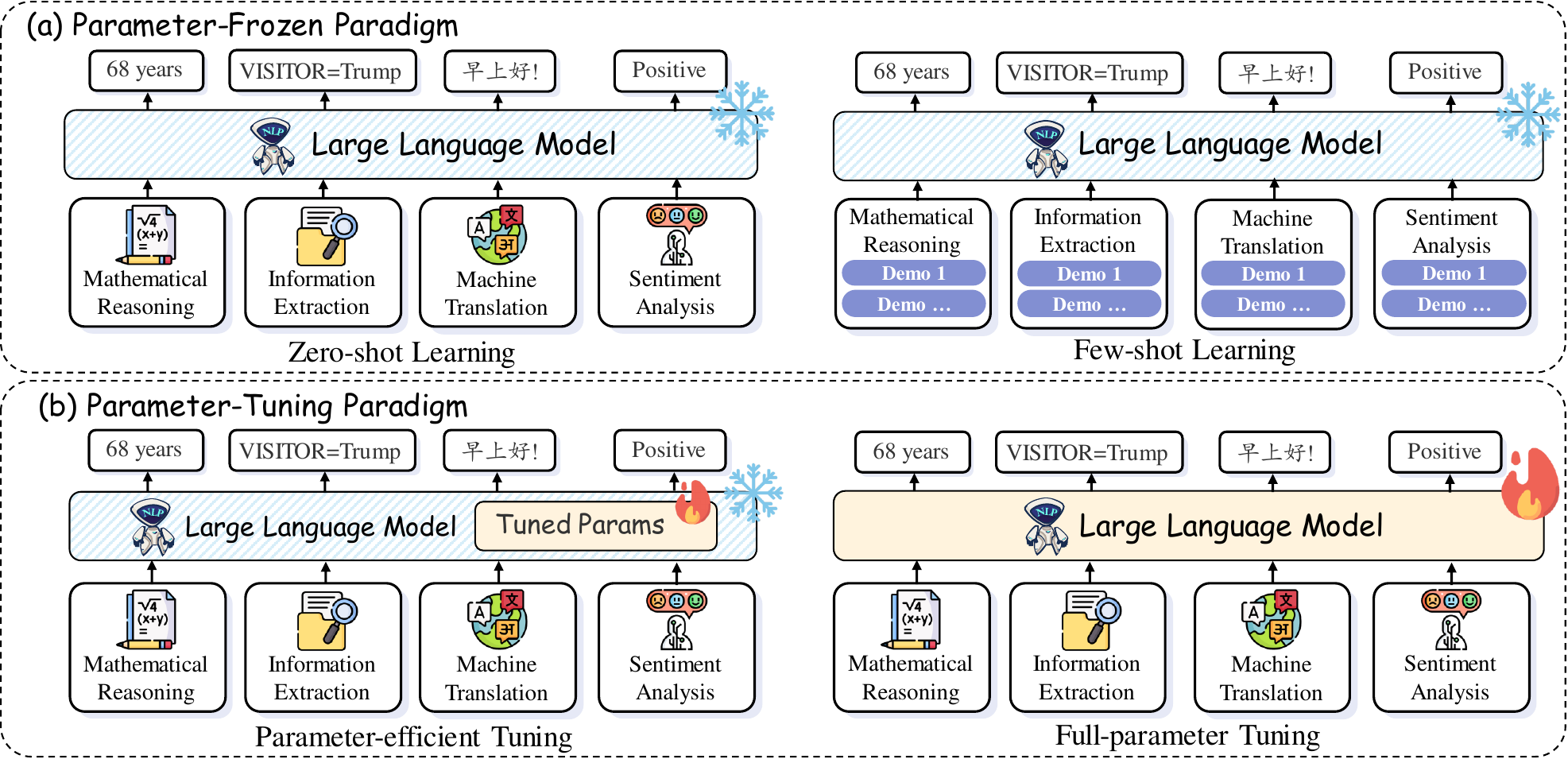}
	\caption{The taxonomy of LLMs for NLP, including parameter-frozen (a) and parameter-tuning paradigm (b), where 
		\fcolorbox{white}{newblue}{ blue module with ice }
		denotes that the parameters are kept unchanged, and
		\fcolorbox{white}{neworange}{ orange module with fire }
		represents the fine-tuning of full or selected parameters.}
	\label{fig:formula}
\end{figure*}

Parameter-frozen paradigm can directly apply prompting for NLP tasks without any parameter tuning. As shown in Figure~\ref{fig:formula} (a), 
this category encompasses \textit{zero-shot learning} and \textit{few-shot learning}~\cite{brown2020language,kojima2022large}.
\paragraph{Zero-shot Learning}
In zero-shot learning, LLMs leverage the instruction following capabilities to solve NLP tasks based on a given instruction prompt, which is defined as:
\begin{equation}
	\mathcal{P} = \texttt{Prompt}(\mathcal{I}),
\end{equation}
where $\mathcal{I}$ and $\mathcal{P}$ denote the input and output of prompting, respectively.

\paragraph{Few-shot Learning}
Few-shot learning uses in-context learning capabilities to solve the NLP tasks imitating few-shot demonstrations.
Formally, given some demonstrations $\mathcal{E}$, the process of few-shot learning is defined as:
\begin{equation}
	\mathcal{P} = \texttt{Prompt}(\mathcal{E}, \mathcal{I}).
\end{equation}
\subsection{Parameter-Tuning Paradigm}\label{sec:parameter-tuning-paradigm}
As shown in Figure~\ref{fig:formula} (b), the parameter-tuning paradigm involves adjusting LLM parameters for NLP tasks, covering both \textit{full-parameter} and \textit{parameter-efficient tuning}.

\paragraph{Full-parameter Tuning}
In the full-parameter tuning approach, all parameters of the model $\mathcal{M}$ are fine-tuned on the training dataset $\mathcal{D}$:
\begin{equation}
	\hat{\mathcal{M}} = \texttt{Fine-tune}(\mathcal{M}|\mathcal{D}),
\end{equation}
where $\hat{\mathcal{M}}$ is {the fine-tuned model with the} updated parameters.

\paragraph{Parameter-efficient Tuning}
Parameter-efficient tuning (PET) involves adjusting a set of existing parameters or incorporating additional tunable parameters (like Bottleneck Adapter~\cite{houlsby2019parameter}, Low-Rank Adaptation {(LoRA)}~\cite{hu2021lora}, Prefix-tuning~\cite{li2021prefix}, {and} QLoRA~\cite{dettmers2023qlora}) to efficiently adapt models for specific NLP tasks.
Formally, parameter-efficient tuning first tunes a set of parameters $\mathcal{W}$, denoting as:
\begin{equation}
	\hat{\mathcal{W}} = \texttt{Fine-tune}(\mathcal{W}|\mathcal{D}, \mathcal{M}),
\end{equation}
where $\hat{\mathcal{W}}$ stands for the trained parameters.

\begin{table*}[t]
\centering
\caption{Comparison of resource consumption and performance across NLP adaptation paradigms. Data is compiled from \citet{mundra2024comprehensive,hu2022lora,dettmers2023qlora}.[+]: better performance / low resource consumption, [++]: much better performance / moderate resource consumption, [+++]: best performance / high resource consumption, [-]: no consumption. Zero-shot Learning has lowest resource consumption and best out-of-domain task generalization, while Full-Parameter Tuning has highest cost and best in-domain performance.}

\resizebox{0.92\textwidth}{!}{
\begin{tabular}{lllcccc}
\toprule
\textbf{Strategy} & \textbf{Training Cost} & \textbf{Memory (Train)} & \textbf{Memory (Infer)} & \textbf{Latency} & \textbf{Accuracy} & \textbf{Generalization} \\
\midrule
Zero-Shot Learning & \$0 & - & + & + & + & +++ \\ 
Few-Shot Learning & \$0 & - & ++ & ++ & ++ & ++ \\ 
\midrule
Full Parameter-Tuning & >\$1K & 2 × model size & + & + & +++ & + \\ 
PET (LoRA) & \$10 $\thicksim $ \$1K & < 1 × model size & + & + & ++ & ++ \\ 
\bottomrule
\end{tabular}
}
\label{tab:resource_comparison}

\end{table*}

\subsection{Comparison of Paradigms}
To further understand the advantages on different paradigms, we summarize the resource consumption and performance of each paradigm in Table~\ref{tab:resource_comparison}.
Generally speaking, zero-shot learning offers the highest application efficiency, moderate improvements on in-domain tasks, and robust out-of-domain generalization. In contrast, few-shot learning typically yields superior in-domain performance relative to zero-shot learning; however, it demands greater computational resources, achieves lower overall efficiency, and exhibits reduced generalization to novel domains. Full-parameter tuning, when ample training data and resources are available, attains the best in-domain performance but at the expense of the least efficient deployment and the weakest transfer to out-of-domain settings. Finally, parameter-efficient tuning strikes a balance: with limited resources, it can match or exceed the performance of full-parameter tuning in certain cases, while offering higher efficiency and often improved generalization beyond the training domain.

\section{Natural Language Understanding}\label{sec:understanding}
As shown in Figure~\ref{fig:taxonomy}, we first describe some typical NLP understanding tasks, which consists of Semantic Analysis~($\S$\ref{sec:SA}), Information Extraction~($\S$\ref{sec:IE}), Dialogue Understanding~($\S$\ref{sec:DU}), and Table Understanding~($\S$\ref{sec:TU}).

\subsection{Sentiment Analysis}
\label{sec:SA}
Sentiment analysis, a key function in natural language processing, identifies the emotional tone of a text, like positive opinions or criticisms~\cite{wankhade2022survey}.
\subsubsection{Parameter-Frozen Paradigm}
\paragraph{Zero-shot Learning}
With the help of instruction tuning, LLMs have been equipped with excellent zero-shot learning ability~\cite{belkhir2023beyond}.
Recent studies~\cite{zhang2023sentiment} find that using simple instructions can elicit ChatGPT's strong capabilities on a series of sentiment analysis tasks such as sentiment classification and aspect-based sentiment analysis.
Current {mainstream} LLMs possess the ability of multilingual understanding to analyze the sentiment conveyed by different languages based on sentiment lexicons~\cite{koto2024zero}. 
Moreover, \citet{du2024evaluation} propose a prompting framework to evaluate and reveal LLMs’ limitations in financial attribute reasoning for sentiment analysis, highlighting weaknesses in numerical and understanding.

\paragraph{Few-shot Learning}
Few-shot prompting not only elicits in-context learning in LLMs but also elaborates the intent of users more clearly. According to the findings presented by previous studies~\cite{zhang2023sentiment, zhao2023chatgpt,xu2023limits, lu2025llm}, incorporating exemplars to the prompts significantly boosts LLMs' performance on aspect-based sentiment analysis and emotion recognition tasks.
Furthermore, \citet{sun2023sentiment} introduce few-shot learning on more complex procedures, incorporating multi-LLM negotiation framework for deeper sentiment analysis.

\tikzstyle{my-box}=[
rectangle,
draw=hidden-black,
rounded corners,
text opacity=1,
minimum height=1.5em,
minimum width=5em,
inner sep=2pt,
align=center,
fill opacity=.5,
]
\tikzstyle{leaf}=[
my-box, 
minimum height=1.5em,
fill=newblue!37, 
text=black,
align=left,
font=\normalsize,
inner xsep=2pt,
inner ysep=4pt,
]
\tikzstyle{leaf2}=[
my-box, 
minimum height=1.5em,
fill=neworange!37, 
text=black,
align=left,
font=\normalsize,
inner xsep=2pt,
inner ysep=4pt,
]
\tikzstyle{space2}=[
my-box, 
minimum height=1.5em,
fill=white, 
text=black,
align=left,
font=\normalsize,
inner xsep=2pt,
inner ysep=4pt,
]
\begin{figure*}[t]
\vspace{-2mm}
\centering
\resizebox{\textwidth}{!}{
\begin{forest}
	forked edges,
	for tree={
		grow=east,
		reversed=true,
		anchor=base west,
		parent anchor=east,
		child anchor=west,
		base=left,
		font=\large,
		rectangle,
		draw=hidden-black,
		rounded corners,
		align=left,
		minimum width=4em,
		edge+={darkgray, line width=1pt},
		s sep=3pt,
		inner xsep=2pt,
		inner ysep=3pt,
		line width=0.8pt,
		ver/.style={rotate=90, child anchor=north, parent anchor=south, anchor=center},
	},
	where level=1{text width=8.4em,font=\normalsize,}{},
	where level=2{text width=11.3em,font=\normalsize,}{},
	where level=3{text width=8.0em,font=\normalsize,}{},
	where level=4{text width=12em,font=\normalsize,}{},
	[Parameter-Frozen Paradigm Taxonomy,ver
	[$\ $Understanding~(\S\ref{sec:understanding}),ver
		[Sentiment Analysis~(\S\ref{sec:SA})
			[\eg~\citet{zhang2023sentiment}{,} \citet{koto2024zero}{,}  \citet{zhao2023chatgpt}{,} \citet{xu2023limits}{,}   \citet{sun2023sentiment}{,}  \citet{du2024evaluation}{,}  \citet{zhang2025revisiting}
				, leaf, text width=48.3em]
		]
		[Information Extraction\\~(\S\ref{sec:IE})
			[\eg~\citet{zhang2023aligning}{,}\citet{wei2023zero}{,}\citet{xie-etal-2023-empirical}{,} \citet{li2023far}{,}\citet{li2023codeie}{,} \citet{bi2023codekgc}{,} \citet{fornasiere2024medical}{,} \\ \citet{tang2024large}
				, leaf, text width=48.3em]
		]
		[Dialogue Understanding\\~(\S\ref{sec:DU})
			[\eg~\citet{pan2023preliminary}{,} \citet{he2023can}{,} \citet{hudevcek2023llms}{,} \citet{heck2023chatgpt}{,}
				\citet{gao2023self}{,}  
				\citet{li2022controllable}{,} \\
				 \citet{zhangseagull,zhang2023sgp}{,} \citet{wu2023semantic}{,}   \citet{das2023s3}{,}  \citet{chi5dialogue}{,} \citet{hu2022context}{,}  \citet{king2023diverse}{,} \citet{addlesee2023multi}{,} \\\citet{chung2023instructtods}{,} \citet{lee2023orchestrallm}{,}  \citet{lin2023toward}{,} \citet{cao2023diaggpt}
				, leaf, text width=48.3em]
		]
		[Table Understanding~(\S\ref{sec:TU})
			[\eg~\citet{singha2023tabular}{,} \citet{patnaik2024cabinet}{,} \citet{ye2023large,ye2024dataframe}{,} \citet{sui2023gpt4table,sui2023tap4llm}{,}  \citet{cheng2022binding}{,}  
			\citet{zhang2023data,zhang-etal-2023-crt,zhang2023reactable,zhang2024large}{,} \\\citet{chen2023large}{,} \citet{luo2023hrot}{,} 
			\citet{li2023sheetcopilot}{,} \citet{jiang-etal-2023-structgpt}{,} \citet{wang2024chain}{,} \citet{kong2024opentab}
			, leaf, text width=48.3em]
		]
	]
	[$\ \ \ $Generation~(\S\ref{sec:generation}),ver
		[Summarization~(\S\ref{sec:SM})
			[\eg~\citet{Goyal2022NewsSA}{,} \citet{Ravaut2023OnCU}{,} \citet{Bhaskar2022PromptedOS}{,} \citet{wang-etal-2023-zero}{,}
				\citet{Zhang2023BenchmarkingLL,Zhang2023ExtractiveSV}{,}  
				\citet{Adams2023FromST}{,} \citet{tang-etal-2023-context}
				, leaf, text width=48.3em]
		]
		[Code Generation~(\S\ref{sec:CG})
			[\eg~\citet{chen2021evaluating}{,} \citet{nijkamp2022codegen}{,} \citet{christopoulou2022pangu}{,} \citet{luo2023wizardcoder}{,} \citet{allal2023santacoder}{,} \citet{li2023starcoder,li2023textbooks}{,} \\
			\citet{guo2024deepseek}{,} \citet{roziere2023code}{,}  \citet{zheng2023codegeex}{,} 
				, leaf, text width=48.3em]
		]
		[Machine Translation~(\S\ref{sec:MT})
			[\eg~\citet{Wei2023PolyLMAO}{,}\citet{Zhu2023ExtrapolatingLL}{,}\citet{li-etal-2023-mt2,Li2023OpenBAAO}{,}\citet{Alves2023SteeringLL}{,}\citet{Raunak2023DissectingIL}{,}\citet{Lu2023ChainofDictionaryPE}
				, leaf, text width=48.3em]
		]
		[Mathematical Reasoning\\~(\S\ref{sec:MR})
			[\eg~\citet{wei2022chain}{,} \citet{zhang2022automatic}{,} \citet{kojima2022large}{,} \citet{wang2023selfconsistency}{,}  \citet{touvron2023llama}{,}   \citet{lu2023dynamic}{,} \citet{gao2023pal}{,} \\
			\citet{das2024mathsensei}
				, leaf, text width=48.3em]
		]]
	]
\end{forest}
}
\resizebox{\textwidth}{!}{
	\begin{forest}
		forked edges,
		for tree={
			grow=east,
			reversed=true,
			anchor=base west,
			parent anchor=east,
			child anchor=west,
			base=left,
			font=\large,
			rectangle,
			draw=hidden-black,
			rounded corners,
			align=left,
			minimum width=4em,
			edge+={darkgray, line width=1pt},
			s sep=3pt,
			inner xsep=2pt,
			inner ysep=3pt,
			line width=0.8pt,
			ver/.style={rotate=90, child anchor=north, parent anchor=south, anchor=center},
		},
		where level=1{text width=8.4em,font=\normalsize,}{},
		where level=2{text width=11.3em,font=\normalsize,}{},
		where level=3{text width=8.0em,font=\normalsize,}{},
		where level=4{text width=12em,font=\normalsize,}{},
		[Parameter-Tuning Paradigm Taxonomy,ver
		[$\ $Understanding~(\S\ref{sec:understanding}),ver
		[Sentiment Analysis~(\S\ref{sec:SA})
		[\eg~\citet{wang2022unifiedabsa}{,} \citet{varia2022instruction}{,} \citet{yang2023visual}{,} \citet{zhao2016sentiment}{,} \citet{qiu2023smile}
		, leaf2, text width=48.3em]
		]
		[Information Extraction\\~(\S\ref{sec:IE})
		[\eg~\citet{lu2023event}{,}\citet{gan2023giellm}{,}\citet{sainz2023gollie}{,}\citet{ wang2023instructuie}{,} \citet{das-etal-2023-unified}{,} \citet{liang-etal-2023-prompts}{,} \citet{dagdelen2024structured}{,} \\
		\citet{xue2024autore}{,} \citet{rixewa2024interleaved}
		, leaf2, text width=48.3em]
		]
		[Dialogue Understanding\\~(\S\ref{sec:DU})
		[\eg~\citet{xie2022unifiedskg}{,}\citet{zhao2022description}{,} \citet{gupta2022show}{,}\citet{yu2022knowledge}{,}\citet{feng2023towards}{,} 
		\citet{liu2023prompt}
		, leaf2, text width=48.3em]
		]
		[Table Understanding~(\S\ref{sec:TU})
		[\eg~\citet{li2023table}{,} \citet{xie2022unifiedskg}{,} \citet{xue2023db}{,} \citet{zhang2023jellyfish}{,} \citet{zhu2024tat}{,}
		\citet{bai2023schema}{,}  \citet{zhang2023tablellama}{,} \\
		\citet{he2025tablelora}{,} \citet{li2024table}
		, leaf2, text width=48.3em]
		]]
		[ $\ \ \ $Generation~(\S\ref{sec:generation}),ver
		[Summarization~(\S\ref{sec:SM})
		[\eg~\citet{Pagnoni2022SocraticPQ}{,} \citet{zhao-etal-2022-domain}{,} \citet{Yuan2022FewshotQS}{,} \citet{Feng2023AdapterbasedSK}{,}
		\citet{Li2021PrefixTuningOC}{,}  \citet{Ravaut2023PromptSumPC}
		, leaf2, text width=48.3em]
		]
		[Code Generation~(\S\ref{sec:CG})
		[\eg~\citet{wang2021codet5,wang2023codet5+}{,}\citet{le2022coderl}{,}\citet{shojaee2023execution}{,}\citet{Ayupov2022ParameterEfficientFO}{,}\citet{zhuo2024astraios}{,}  \citet{weyssow2023exploring}
		, leaf2, text width=48.3em]
		]
		[Machine Translation~(\S\ref{sec:MT})
		[\eg~\citet{Xu2023APS,Xu2024ContrastivePO}{,} \citet{Iyer2023TowardsED}{,}\citet{Moslem2023FinetuningLL}{,}\citet{Ustun2022WhenDP}{,}\citet{Alves2023SteeringLL} \citet{Wu2023ExtrapolatingMU,Wu2024AdaptingLL}{,}
		, leaf2, text width=48.3em]
		]
		[Mathematical Reasoning\\~(\S\ref{sec:MR})
		[\eg~~\citet{luo2023wizardmath}{,} \citet{yue2023mammoth}{,} \citet{ho-etal-2023-large}{,} \citet{schick2023toolformer}{,} \citet{hu2022lora,hu2023llmadapters}{,} \citet{shi2024math}{,} \citet{shao2024deepseekmath}{,}  \\
		\citet{luo2024improve}{,} \citet{chen2024masked}{,} 
		\citet{liu2023plan}{,} \citet{yu2024reasonagain}{,} \citet{ranaldi2025improving}{,} \citet{srivastava2025debate}{,} \citet{cai2024system}
		, leaf2, text width=48.3em]
		]
		]
		]
	\end{forest}
}
\vspace{-4mm}
\caption{Taxonomy of LLMs for NLP including Parameter-Frozen Paradigm and Parameter-Tuning Paradigm.
}
\label{fig:taxonomy}
\vspace{-3mm}
\end{figure*}
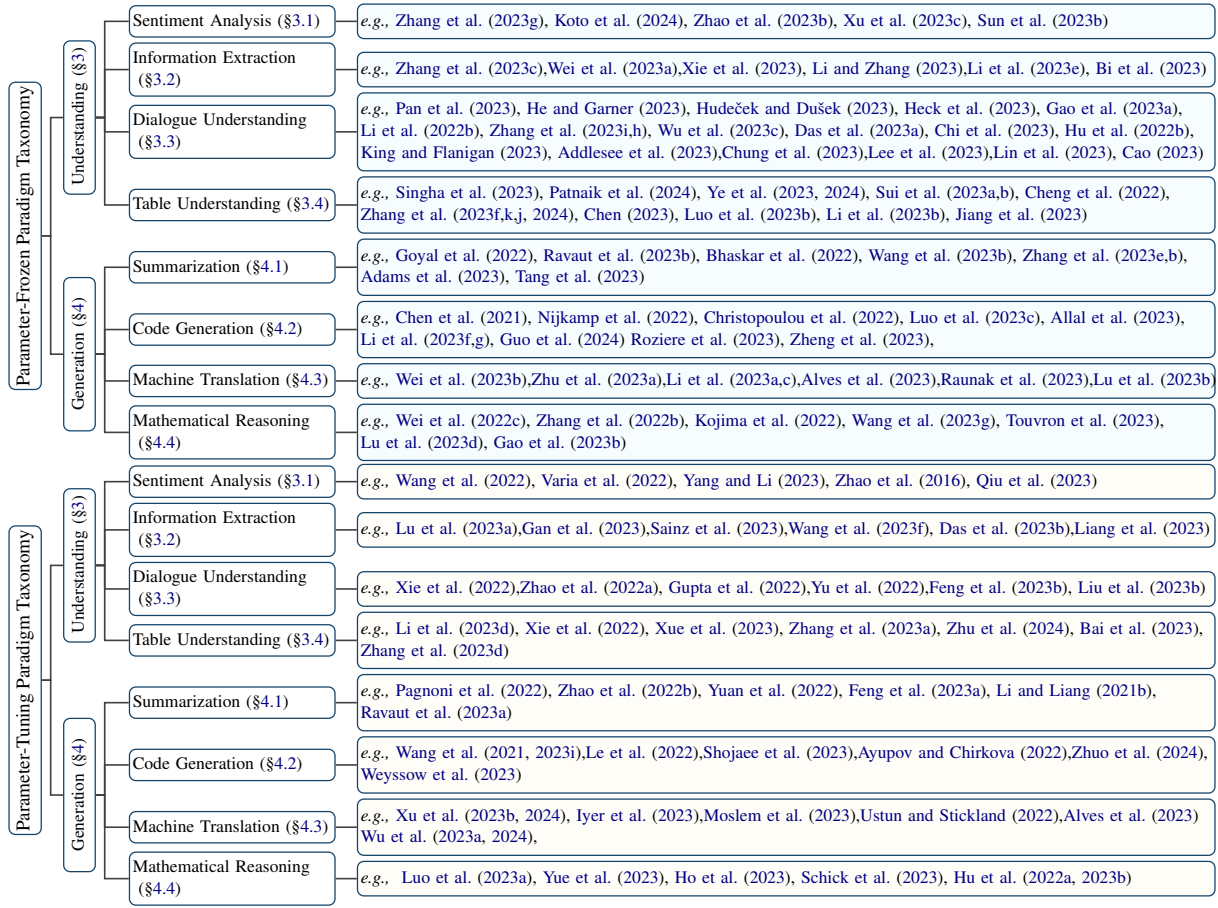

\subsubsection{Parameter-Tuning Paradigm}
\paragraph{Full-Parameter Tuning}
Full-parameter instruction tuning has been shown to be an effective approach to bridge the gap between task-agnostic pre-training and task-specific inference.
Specifically, \citet{wang2022unifiedabsa} design unified sentiment instruction for various aspect-based sentiment analysis tasks to elicit the LLMs. \citet{varia2022instruction} utilize task-specific sentiment instructions to fine-tune LLMs for the inter-task dependency. \citet{yang2023visual} transform the visual input into plain text during prompt construction for instruction tuning. 
Moreover, \citet{zhang2025revisiting} conduct an empirical study to evaluate bLLMs' effectiveness in sentiment analysis for software engineering, revealing their advantages in data-scarce scenarios and limitations compared to fine-tuned sLLMs with sufficient training data.
These works demonstrate the potential of tuning LLMs for advanced sentiment analysis. 

\paragraph{Parameter-Efficient Tuning}
Sentiment analysis techniques have numerous real-world applications such as opinion mining~\cite{zhao2016sentiment}. Therefore, efficiency is a vital dimension for evaluating sentiment analysis methods. 
\citet{qiu2023smile} utilize LoRA to tune LLMs on the empathy multi-turn conversation dataset namely SMILECHAT to develop emotional systems.
\subsection{Information Extraction}
\label{sec:IE}
Information Extraction (IE) tasks aim at extracting structural information from plain text, which typically includes relation extraction (RE), named entity recognition (NER), and event extraction (EE)~\cite{xu2023large}.
\subsubsection{Parameter-Frozen Paradigm}
\paragraph{Zero-shot Learning}
Inspired by the impressive capabilities of LLMs on various tasks, recent studies~\cite{zhang2023aligning,wei2023zero} begin to explore zero-shot prompting methods to solve IE tasks by leveraging knowledge embedded in LLMs.
\citet{wei2023zero} {,} \citet{xie-etal-2023-empirical} {and} \citet{zhang2023aligning} propose a series of methods to decompose question-answering tasks by breaking down NER into smaller, simpler subproblems, which improves the overall process.
\citet{xie-etal-2023-empirical} introduce two methods, syntactic prompting and tool augmentation, to improve performance of LLMs by incorporating the syntactic information.
\citet{siepmann2025automated} explore the use of GPT-4 for automated information extraction of diagnoses, medications, and allergies from discharge letters, demonstrating high accuracy with prompt tuning and highlighting its potential to reduce administrative burden in healthcare.
In addition, \citet{gu2025scalable} conduct a cross-sectional study to evaluate advanced open-source LLMs for information extraction of social determinants of health (SDoH) from clinical notes, using human-annotated EHR data and comparing against a pattern-matching baseline.

\paragraph{Few-shot Learning}
Considering the gap between sequence labeling and text generation, providing exemplars could help LLMs better understand the given task and follow the problem-solving steps, especially in tasks requiring structured outputs and clear format adherence for accuracy.
To select pertinent demonstrations, \citet{li2023far} deploy the retrieval module to retrieve the most suitable examples for the given test sentence, aiming to enhance task relevance and response accuracy.
Instead of using natural language for structured output, \citet{li2023codeie} and \citet{bi2023codekgc} propose reformulating IE tasks as code with code-related LLMs such as Codex, effectively leveraging their powerful syntax-aware generation and reasoning capabilities. \citet{fornasiere2024medical} introduce prompt-based strategies for small-scale LLMs to extract structured and unstructured medical information from clinical texts, demonstrating strong zero-shot performance and enhanced explainability through line-number referencing to source text. \citet{tang2024large} explore the impact of various prompt engineering strategies, persona, chain-of-thought, and few-shot prompting, on the performance of GPT-3.5 and GPT-4 in extracting key information from medical publications, evaluating alignment with ground truth using multiple comprehensive metrics.
\subsubsection{Parameter-Tuning Paradigm}
\paragraph{Full-Parameter Tuning}
A common practice to customize LLMs is fine-tuning LLMs on the collected dataset. There typically are three tuning paradigms adopted to enhance LLMs' abilities. The first one is tuning LLMs on a single dataset to strengthen a specific ability. 
The second one is standardizing data formats across all IE subtasks, thus enabling a single model to efficiently handle diverse tasks~\citep{lu2023event,gan2023giellm}. The last one is tuning LLMs on a mixed dataset and testing on the unseen tasks~\cite{sainz2023gollie, wang2023instructuie}, which is always used to improve the generalization ability of LLMs. \citet{rixewa2024interleaved} introduce a unified interleaved representation with cross-modal attention to enhance multi-modal information retrieval, enabling accurate and efficient processing of complex content across text and image formats.

\paragraph{Parameter-Efficient Tuning}
Tuning huge parameters of LLMs poses a significant challenge to both research and development. To address this challenge \cite{xin2024parameter,xin2024v}, \citet{das-etal-2023-unified}
propose a method for dynamic sparse fine-tuning that focuses on a specific subset of parameters during the IE training process. This approach is particularly useful when dealing with limited data. 
Meanwhile, \citet{liang-etal-2023-prompts}
introduce Lottery Prompt Tuning (LPT), a method that efficiently tunes only a portion of the prompt vectors used for lifelong information extraction. This technique optimizes both parameter efficiency and deployment efficiency. \citet{dagdelen2024structured} introduce a simple and flexible approach to fine-tuning LLMs for joint named entity recognition and relation extraction, enabling the generation of structured scientific knowledge records from complex materials chemistry texts. \citet{xue2024autore} introduce AutoRE, a novel end-to-end document-level relation extraction model using the RHF paradigm and parameter-efficient fine-tuning, enabling state-of-the-art performance without relying on predefined options.
\subsection{Dialogue Understanding}
\label{sec:DU}

Dialogue understanding typically consists of spoken language understanding (SLU)~\cite{ijcai2021p622} and dialogue state tracking (DST)~\cite{sarikaya2016overview}.
\subsubsection{Parameter-Frozen Paradigm}
\paragraph{Zero-shot Learning}
Recent studies highlight the effectiveness of LLMs in dialogue understanding through zero-shot prompting~\cite{pan2023preliminary,he2023can,hudevcek2023llms,heck2023chatgpt,yoon2024blendx}.
\citet{gao2023self} and \citet{addlesee2023multi} introduce zero-shot chain-of-thought prompting strategies in LLMs, enhancing understanding by step-by-step reasoning.
Moreover, \citet{zhangseagull} and \citet{wu2023semantic} treat SLU and DST as agent systems and code generation tasks to effectively improve task performance.
Further, \citet{chung2023instructtods}, \citet{chi5dialogue} and \citet{zhang2023sgp} extend the task to actual scenarios and understand the dialog by zero-shot prompting for efficient interaction and dialog management.
Recently, \citet{qin2025divide} and \citet{qin2025croprompt} propose a series of multi-stage solution frameworks that leverages the interactive capabilities of LLMs to address single‑intent and multi‑intent SLU tasks respectively.
\citet{dong-etal-2025-protod} propose a multi-agent framework, ProTOD, which is a novel active DST planner framework based on multiple LLMs' interation, designed to enhance the dialog's proactivity and goal completion rate.

\paragraph{Few-shot Learning}

Limited by the instruction following ability of the LLMs, recent studies have focused on improving model performance in dialogue understanding through the relevant few-shot demonstrations~\cite{hudevcek2023llms}. 
To address ``overfitting'' in the given few-shot demonstrations, \citet{hu2022context}, \citet{king2023diverse}, \citet{das2023s3}, \citet{li2022controllable}, \citet{lee2023orchestrallm}, \citet{king2023diverse} and \citet{addlesee2023multi} further introduce some methods for retrieving diverse few-shot demonstrations to improve understanding performance.
\citet{lin2023toward} and \citet{cao2023diaggpt} integrate DST tasks with an agent through in-context-learning, enhancing dialogue understanding capabilities.

\subsubsection{Parameter-Tuning Paradigm}

\paragraph{Full-Parameter Tuning} Full-parameter tuning involves not freezing any parameters and using all parameters to train dialogue understanding tasks~\cite{yu2022knowledge}. Specifically, \citet{xie2022unifiedskg,zhao2022description} unifies structured tasks into a textual format by training full parameters demonstrating significant improvement and generalization.
\citet{gupta2022show} utilize input with some demonstrations as a new DST representation format to train LLM with full parameters and achieve great results.
\citet{acikgoz2025can} suggest that DST, typically trained on a limited set of APIs, needs new data for quality maintenance. They propose a unified instruction-tuning paradigm for multi-turn DST and advanced function calls, enhancing dialogue management and generalization.
\paragraph{Parameter-Efficient Tuning}
Limited by the huge cost of full-parameter fine-tuning, a lot of work begins to focus more on Parameter-Efficient Tuning (PET) for lower-cost dialogue understanding task training.
Specifically, \citet{feng2023towards} present LDST, a LLaMA-driven DST framework that leverages LoRA technology for parameter-efficient fine-tuning, achieving performance comparable to ChatGPT.
\citet{liu2023prompt} provide a key-value pair soft-prompt pool, selecting soft-prompts from the prompting pool based on the conversation history for better PET. Further \citet{yin-etal-2025-midlm} address the multi-intent detection task and introduces MIDLM, a bidirectional LLM framework that enables autoregressive LLMs to leverage bidirectional information through post-training, thereby eliminating the need to train the model from scratch.

\subsection{Table Understanding}
\label{sec:TU}
Table understanding involves the comprehension and analysis of structured data presented in tables, focusing on interpreting and extracting meaningful information, like Table Question Answering~\cite{jin2022survey,wang2022survey,zhang2025survey}.
\subsubsection{Parameter-Frozen Paradigm}
\paragraph{Zero-shot Learning} 
Recently, the advancements for LLMs have paved the way for exploring zero-shot learning capabilities in understanding and interpreting tabular data~\cite{singha2023tabular,patnaik2024cabinet,ye2024dataframe}.
\citet{ye2023large} and \citet{sui2023gpt4table} concentrate on breaking down large tables into smaller segments to reduce irrelevant data interference during table understanding.
Further, \citet{patnaik2024cabinet} introduce CABINET, a framework that includes a module for generating parsing statements to emphasize the data related to a given question.
\citet{sui2023tap4llm} develop TAP4LLM, enhancing LLMs' table understanding abilities by incorporating reliable information from external knowledge sources into prompts. 
Additionally, \citet{ye2024dataframe} propose a DataFrameQA framework to utilize secure Pandas queries to address issues of data leakage in table understanding.
These efforts signify a significant stride towards leveraging LLMs for more effective and efficient zero-shot learning in table data comprehension.

\paragraph{Few-shot Learning}
Few-shot learning has been an increasingly focal point for researchers to address the limitations of LLMs, particularly in the context of table understanding and instruction following ability~\cite{chen2023large,zhang2024large,zhang2025rot}.
\citet{luo2023hrot} propose a hybrid prompt strategy coupled with a retrieval-of-thought to further improve the example quality for table understanding tasks.
\citet{cheng2022binding} introduce Binder to redefine the table understanding task as a coding task, enabling the execution of code to derive answers directly from tables.
Furthermore, \citet{li2023sheetcopilot}, \citet{jiang-etal-2023-structgpt} and \citet{zhang-etal-2023-crt,zhang2023data} conceptualize the table understanding as a more complex agent task, which utilizes external tools to augment LLMs in table tasks.
Building upon these developments, ReAcTable~\cite{zhang2023reactable} integrates additional actions into the process, such as generating SQL queries, producing Python code, and directly answering questions, thereby further enriching the few-shot learning landscape. \citet{wang2024chain} introduce Chain-of-Table, a framework that guides LLMs to perform table-based reasoning by iteratively updating tabular data as intermediate steps, enabling structured, dynamic reasoning chains that significantly improve performance on table understanding tasks. \citet{kong2024opentab} propose OpenTab, an open-domain table reasoning framework that enhances LLMs' ability to handle structured table data by retrieving relevant tables and generating SQL programs for reasoning, significantly improving accuracy over existing methods in both open and closed domain scenarios.

\subsubsection{Parameter-Tuning Paradigm}
\paragraph{Full-Parameter Tuning} 
Leveraging the existing capabilities of LLMs, Full-Parameter Tuning optimizes these models for specific table understanding tasks.
\citet{li2023table} and \citet{xie2022unifiedskg} adapt a substantial volume of table-related data for table instruction tuning, which leads to better generalization in table understanding tasks. 
Additionally, \citet{xue2023db} introduce DB-GPT to enhance LLMs by fine-tuning them and integrating a retrieval-augmented generation component to better support table understanding.
\paragraph{Parameter-Efficient Tuning}
\citet{xie2022unifiedskg} utilize prompt-tuning for efficient fine-tuning within a unified framework of table representation instructions.
Moreover, \citet{zhang2023jellyfish}, \citet{zhu2024tat} and \citet{bai2023schema} adapt Low-Rank Adaptation (LoRA) during instruction-tuning for better table understanding and further table cleaning.
Furthermore, \citet{zhang2023tablellama} address challenges related to long table inputs by implementing LongLoRA, demonstrating its efficacy in managing long-context issues in table understanding tasks. \citet{he2025tablelora} introduce TableLoRA, a table-specific fine-tuning module that enhances LLMs’ understanding of tabular data under parameter-efficient settings by combining specialized table serialization and 2D positional encoding to improve performance on structured table tasks. \citet{li2024table} introduce a new ``table fine-tuning'' paradigm that enhances language models like GPT-3.5 and ChatGPT on diverse table-understanding tasks by training them with synthesized table-based instructions, significantly improving their performance and generalizability on structured tabular data.

\section{Natural Language Generation}
\label{sec:generation}
This section presents the LLMs for classific NLP generation tasks containing Summarization~($\S$\ref{sec:SM}), Code Generation~($\S$\ref{sec:CG}), Machine Translation~($\S$\ref{sec:MT}), and Mathematical Reasoning~($\S$\ref{sec:MR}), which are illustrated in Figure~\ref{fig:taxonomy}.

\subsection{Summarization}\label{sec:SM}

Summarization aims to distill the essential information from a text document, producing a concise and coherent synopsis that retains the original content's themes~\citep{Shi2018NeuralAT}.

\subsubsection{Parameter-Frozen Paradigm}
\paragraph{Zero-shot Learning}
In the exploration of zero-shot learning for text summarization, LLMs such as GPT-3 have demonstrated amazing and superior performance in generating concise and factually accurate summaries, challenging the need for traditional fine-tuning approaches \citep{Goyal2022NewsSA,Bhaskar2022PromptedOS,wang-etal-2023-zero}. 
\citet{Zhang2023BenchmarkingLL} highlight instruction tuning as pivotal for LLMs' summarization success.
\citet{Ravaut2023OnCU} scrutinize LLMs' context utilization, identifying a bias towards initial document segments in summarization tasks~\citep{godbole2025leveraging,peters2025generalization}.
Furthermore, \citet{yun2025summpilot} enhances automatic summarization by integrating human interaction and semantic graphs, enabling the generation of higher-quality, personalized summaries tailored to individual users' interests and needs.
These studies collectively underscore the versatility and challenges of deploying LLMs in zero-shot summarization.

\paragraph{Few-shot Learning}
For few-shot learning, LLMs like ChatGPT are scrutinized for their summarization abilities. 
\citet{Zhang2023ExtractiveSV} and \citet{tang-etal-2023-context} demonstrate that leveraging in-context learning and a dialog-like approach can enhance LLMs' extractive summarization, particularly in achieving summary faithfulness. 
\citet{Adams2023FromST} introduce a ``Chain of Density” prompting technique, revealing a preference for denser, entity-rich summaries over sparser ones. 
Moreover, recent studies have begun to leverage the reflective capabilities~\citep{qorib2025just}, deeper reasoning abilities~\citep{zhu2025understanding}, and planning abilities~\citep{nandy2025language} of large reasoning models to enhance the depth of thought as well as the conciseness and clarity of summaries.
Together, these studies reveal the evolving strategies to optimize LLMs for summarization tasks.

\subsubsection{Parameter-Tuning Paradigm}
\paragraph{Full-Parameter Tuning}
Full-Parameter Tuning for text summarization leverages the power of LLMs, optimizing them for specific summarization tasks.  
DIONYSUS~\citep{Li2022DIONYSUSAP} adapts to new domains through a novel pre-training strategy tailored for dialogue summarization. 
Socratic Pretraining~\citep{Pagnoni2022SocraticPQ} introduces a question-driven approach to improve the summarization process.
Further, \citet{wang2025distilling} and \citet{lu2025mutual} demonstrate that carefully prompting LLMs produces well‑structured rationales, which can guide smaller models with fully tuning to generate summaries that are both more concise and of higher quality.
More recently, \citet{aali2025dataset} and \citet{wu2025rlpf} employ meticulously annotated supervised fine-tuning (SFT) data and prediction feedback-based reinforcement learning, respectively, enabling their models to match or even surpass the performance of proprietary closed-source models.
Overall, this allows the model to be easily adapted for different summarization tasks, resulting in more controllable and relevant summaries.

\paragraph{Parameter-Efficient Tuning}
PET strategies have revolutionized the adaptability of large pre-trained models for specific summarization tasks, demonstrating the power of fine-tuning with minimal parameter adjustments~\cite{Feng2023AdapterbasedSK}. 
\citet{zhao-etal-2022-domain} and \citet{Yuan2022FewshotQS} adapt prefix-tuning~\citep{Li2021PrefixTuningOC} for dialogue summarization, enhancing model knowledge and generalization across domains. 
\citet{Ravaut2023PromptSumPC} develop PromptSum to combine prompt tuning with discrete entity prompts for controllable abstractive summarization.
These approaches collectively show the efficacy of PET in enabling robust, domain-adaptive, and controllable summarization with minimal additional computational costs.
\subsection{Code Generation}\label{sec:CG}

Code generation involves the automatic creation of executable code from natural language specifications, facilitating a more intuitive interface for programming~\citep{chen2021evaluating}.

\subsubsection{Parameter-Frozen Paradigm}

\paragraph{Zero-shot Learning}
Recent advancements in code generation have been significantly propelled by the development of LLMs, with studies showcasing their proficiency in generating code in a zero-shot manner. 
Code LLMs, trained on both code and natural language, have a robust and amazing zero-shot learning capability for programming tasks~\citep{nijkamp2022codegen,roziere2023code}. 
Moreover, CodeT5+ enriches the landscape by proposing a flexible encoder-decoder architecture and a suite of pretraining objectives, leading to notable improvements \citep{wang2023codet5+}. 
These models collectively push the boundary of what is achievable in code generation, offering promising avenues for zero-shot learning.
Recent releases of code-specific LLMs, such as CodeGemma~\citep{team2024codegemma} and Qwen2.5-Coder~\citep{hui2024qwen2}, further advance the field of LLM-based code generation, delivering superior benchmark performance. 
Additionally, Seed-Coder~\citep{seed2025coder} introduces a model-centric data curation pipeline, while Ling-Coder-Lite~\citep{team2025every} leverages a Mixture-of-Experts architecture to balance efficiency and performance, marking state-of-the-art progress in open-source code generation LLMs.

\paragraph{Few-shot Learning}
Code generation is being revolutionized by few-shot learning. This technique allows models to create precise code snippets by learning from just minimal examples~\citep{lu2021codexglue}.
\citet{chen2021evaluating}, \citet{allal2023santacoder}, \citet{li2023starcoder}, \citet{luo2023wizardcoder}, and \citet{christopoulou2022pangu} illustrate the efficacy of few-shot learning, demonstrating an adeptness at code generation that surpasses its predecessors. 
The development of smaller, yet powerful models~\citep{li2023textbooks,guo2024deepseek} further highlights the accessibility of few-shot code generation technologies, making them indispensable tools in the arsenal of modern developers. 
Importantly, most modern LLMs for code generation, including Code Llama~\citep{roziere2023code}, Seed-Coder~\citep{seed2025coder}, Qwen2.5-Coder~\citep{hui2024qwen2}, and CodeGemma~\citep{team2024codegemma}, provide both base and instruct variants, enabling flexible few-shot learning execution across diverse programming tasks.

\subsubsection{Parameter-Tuning Paradigm}

\paragraph{Full-Parameter Tuning}
Full-parameter tuning represents a pivotal strategy in enhancing code generation models, allowing comprehensive model optimization.
Specifically, CodeT series~\citep{wang2021codet5,wang2023codet5+} epitomize this approach by incorporating code-specific pre-training tasks and architecture flexibility, respectively, to excel in both code understanding and generation. 
CodeRL~\citep{le2022coderl} and PPOCoder~\citep{shojaee2023execution} introduce deep reinforcement learning, leveraging compiler feedback and execution-based strategies for model refinement, whereas StepCoder~\citep{shojaee2023execution} advances this further by employing reinforcement learning, curriculum learning, and fine-grained optimization techniques. 
These models collectively demonstrate significant improvements across a spectrum of code-related tasks, embodying the evolution of AI-driven programming aids.
Emerging work such as PRLCoder~\citep{ye2025process} leverages process-supervised reinforcement learning, Focused-DPO~\citep{zhang2025focused} enhances preference optimization on error-prone points, and ACECoder~\citep{zeng2025acecoder} applies automated test-case synthesis to refine reward models. 
Furthermore, SWE-RL~\citep{wei2025swe} expands reinforcement learning into real-world software engineering, significantly advancing the reasoning capacities of LLMs. 
Reinforcement learning thus demonstrates strong potential for training code LLMs and warrants further exploration.

\paragraph{Parameter-Efficient Tuning}
PET emerges as a pivotal adaptation in code tasks, striking a balance between performance and computational efficiency~\citep{weyssow2023exploring}. 
Studies~\citep{Ayupov2022ParameterEfficientFO,zhuo2024astraios} exploring adapters and LoRA showcase PET's viability on code understanding and generation tasks, albeit with limitations in performance.
Recent investigations, such as~\citet{storhaug2024parameter}, demonstrate that PEFT methods can rival full fine-tuning for unit test generation, reducing resource demands. Additionally,~\citet{zhang2024comprehensive} provide a comprehensive evaluation of PEFT on method-level code smell detection, revealing that small models often perform competitively, reinforcing the scalability and cost-effectiveness of PET techniques for specialized software engineering tasks.

\subsection{Machine Translation}\label{sec:MT}

Machine translation is a classical task that utilizes computers to automatically translate the given information from one language to another, striving for accuracy and preserving the semantic essence of the original material~\citep{Bahdanau2014NeuralMT}. 
Recent work~\citep{pang2025salute} revisits key challenges in neural machine translation (NMT), highlighting how LLMs address issues such as long sentence translation and reduced parallel data reliance while facing new challenges like inference efficiency and low-resource language translation.

\subsubsection{Parameter-Frozen Paradigm}

\paragraph{Zero-shot Learning}
In the realm of zero-shot learning, 
\citet{Zhu2023ExtrapolatingLL} and \citet{Wei2023PolyLMAO} enhance LLMs' multilingual performance through cross-lingual and multilingual instruction-tuning, significantly improving translation tasks. OpenBA contributes to the bilingual model space, demonstrating superior performance in Chinese-oriented tasks with a novel architecture~\citep{Li2023OpenBAAO}. These advancements highlight the potential of LLMs in aligning language in zero-shot settings.

\paragraph{Few-shot Learning}
In the exploration of few-shot learning for machine translation (MT), recent studies present innovative strategies to enhance the capabilities of LLMs~\cite{li-etal-2023-mt2,Huang2024AligningTU}. 
\citet{Lu2023ChainofDictionaryPE} introduce Chain-of-Dictionary Prompting (CoD) to improve the MT of rare words by in-context learning in low-resource languages. 
\citet{Raunak2023DissectingIL} investigate the impact of demonstration attributes on in-context learning, revealing the critical role of output text distribution in translation quality. 
\citet{zhu2024towards} propose a robust multi-view approach for selecting fine-grained demonstrations, effectively reducing noise in in-context learning and significantly improving domain adaptation. Together, these works illustrate the significant potential of few-shot learning in advancing the field of MT with LLMs.

\subsubsection{Parameter-Tuning Paradigm}
\paragraph{Full-Parameter Tuning}
Full-parameter tuning in machine translation with LLMs represents a frontier for enhancing translation accuracy and adaptability~\cite{Xu2023APS}. 
\citet{Iyer2023TowardsED} demonstrate the potential of LLMs in disambiguating polysemous words through in-context learning and fine-tuning on ambiguous datasets, achieving superior performance in multiple languages.  
\citet{Moslem2023FinetuningLL} and \citet{Wu2024AdaptingLL} focus on exploring fine-tuning methods that enhance real-time and context-aware translation capabilities. 
\citet{Xu2024ContrastivePO} propose Contrastive Preference Optimization (CPO) to refine translation quality further, pushing LLMs towards better performance. 
\citet{feng2025mt} introduce MT-R1-Zero, applying reinforcement learning frameworks to MT without supervised fine-tuning, achieving competitive results on multilingual benchmarks and offering insights into emergent reasoning patterns. 
\citet{feng2024ladder} present MT-Ladder, a cost-effective hierarchical fine-tuning framework that boosts general-purpose LLMs' translation performance to match state-of-the-art models. 
These studies reveal the efficacy and necessity of fine-tuning approaches, and point toward reinforcement learning as a promising future direction for advancing machine translation by leveraging LLMs' emergent reasoning and adaptability.

\paragraph{Parameter-Efficient Tuning}
PET is emerging as a transformative approach for integrating LLMs into machine translation (MT), balancing performance and efficiency. 
\citet{Ustun2022WhenDP} empirically assess PET's efficacy across different languages and model sizes, highlighting adapters' effectiveness with adequate parameter budgets. 
\citet{Alves2023SteeringLL} optimize the fine-tuning process with adapters, striking a balance between few-shot learning and fine-tuning efficiency. 
Recent work further demonstrates PET's scalability and robustness in multilingual and domain-specific tasks, confirming its potential to make LLMs more adaptable and resource-efficient while maintaining competitive performance. These studies collectively underline PET's promise to revolutionize MT by offering scalable and cost-effective solutions.

\subsection{Mathematical Reasoning}
\label{sec:MR}
Mathematical reasoning tasks in NLP involve the use of NLP techniques to understand information from mathematical text, perform logical reasoning processes, and ultimately generate accurate answers to mathematical questions~\cite{lu2023survey,yan2025mathagent}.
\subsubsection{Parameter-Frozen Paradigm}
\paragraph{Zero-shot Learning}
Mathematics serves as a testbed to investigate the reasoning capabilities of LLMs~\cite{openai2023gpt4,touvron2023llama}. The vanilla prompting method asks LLMs to directly arrive at the final answer to a given mathematical problem. It is very challenging and the reasoning process is not transparent to humans.
To address it, \citet{kojima2022large} develop a zero-shot chain-of-thought technique, which utilizes the simple prompt ``Let's think step by step'' to elicit mathematical reasoning in LLMs.
By doing this, the LLM can break down the problem into smaller, easier-to-solve pieces before arriving at a final answer.
Further, \citet{wang2023selfconsistency} propose a new decoding strategy, called self-consistency. This approach integrates a series of prompting results to boost the performance.
\citet{tang2025elevating} propose an automatically enhanced zero-shot prompting strategy that adjusts the prompts through model retrieval to improve the performance of LLMs on mathematical reasoning tasks.
Moreover, \citet{yuksekgonul2025optimizing} and \citet{peng2025dlpo} employ reflection-based, iterative prompting strategies to improve zero-shot mathematical reasoning accuracy.

\paragraph{Few-shot Learning}
Recent studies explore constructing more suitable exemplars for LLMs to improve mathematical reasoning.  \citet{wei2022chain} introduce chain-of-thought prompting, using a few demonstrations to guide LLMs through step-by-step reasoning. However, creating these examples by hand is laborious, so \citet{zhang2022automatic} and \citet{lu2023dynamic} propose methods to select in-context examples automatically. To improve numerical precision, PAL~\cite{gao2023pal} generates and executes intermediate program steps in a runtime environment. Building on this idea, \citet{das2024mathsensei} present MathSensei, a tool-augmented LLM that integrates web search, code execution, and symbolic solving, showing greater gains on harder problems. \citet{liu2023plan} propose XoT, a unified framework that dynamically switches among diverse prompting methods for better math reasoning. To probe consistency, \citet{yu2024reasonagain} use symbolic programs to reveal that LLMs often rely on brittle reasoning despite strong static performance.
More recently, \citet{ranaldi2025improving} introduce QuaSAR, which blends natural language with selective formalization to enhance chain-of-thought robustness without full symbolic translation. Moreover, \citet{zhang2025booststep} enhance the mathematical capabilities of LLMs by improving their single-step reasoning in the context of fine-grained in-context learning.

\begin{figure*}[t]
	\centering
	\includegraphics[width=0.86\textwidth]{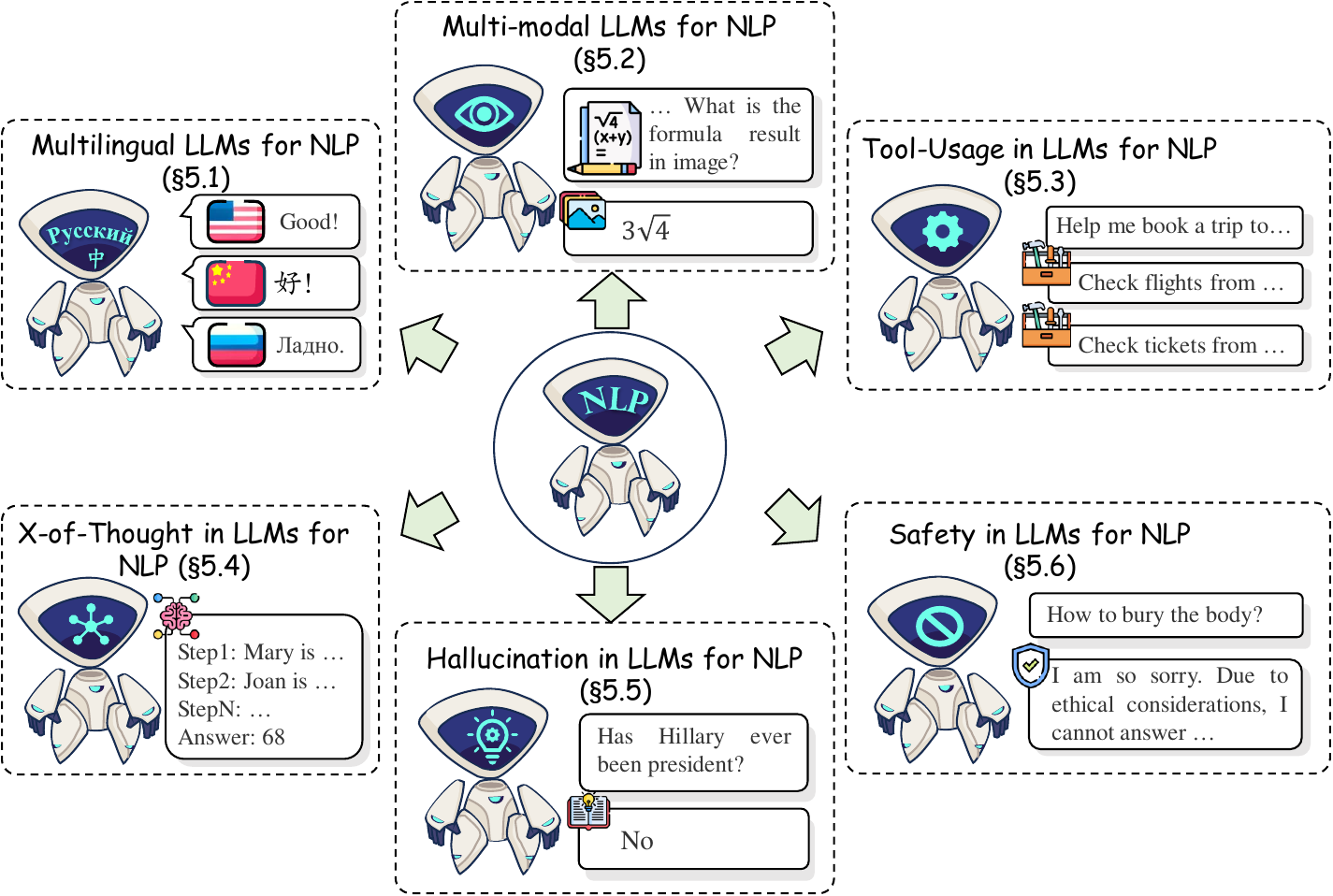}
	\caption{The future work and new frontier for LLM in NLP tasks.}
	\label{fig:future}
\end{figure*}

\subsubsection{Parameter-Tuning Paradigm}
\paragraph{Full-Parameter Tuning}
Full-parameter tuning is a standard method for guiding LLMs in mathematical reasoning tasks~\citep{cai2024system}. Several studies have improved general math-solving ability by creating high-quality instruction-tuning datasets, from web-curated collections~\citep{yue2023mammoth}, advanced LLM distillation~\citep{ho-etal-2023-large}, and self-generated samples~\citep{pang2025bolt,cai2024system}.
Moreover, \citet{schick2023toolformer} introduce ToolFormer, which leverages a calculator for numeric operations. \citet{chen2024masked} propose perturbing token-level chain-of-thought during fine-tuning, improving accuracy without external labels. \citet{yu2025chain} develop Chain-of-Reasoning, which integrates natural-language, algorithmic, and symbolic reasoning to boost benchmarks.
Beyond supervised tuning, reinforcement learning has also shown promise. \citet{luo2023wizardmath} apply RLEIF to enhance math reasoning; \citet{luo2024improve} propose OmegaPRM, an MCTS-based method for training reward models on MATH500 and GSM8K without human oversight; \citet{shao2024deepseekmath} train DeepSeekMath 7B on 120 B tokens using web data and GRPO; and \citet{qian2025toolrl} introduce ToolRL, examining tool selection and reward design in RL-based fine-tuning.

\paragraph{Parameter-Efficient Tuning}
Fine-tuning LLMs with full parameter updates incurs significant memory overhead, limiting accessibility for many users. Parameter-efficient tuning techniques, such as LoRA~\cite{hu2022lora}, offer a promising alternative.
Additionally, \citet{hu2023llmadapters} propose a user-friendly framework for integrating various adapters into LLMs, enabling them to tackle tasks like mathematical reasoning.
SPHERE~\citep{singh2025self} introduces a self‑evolving data‑generation pipeline leveraging LoRA to enhance the performance of small‑scale language models on mathematical reasoning tasks through the self‑generation, refinement, and diversification of reasoning chains.
\citet{prottasha2024parameter} present Semantic Knowledge Tuning (SK‑Tuning), which employs semantically meaningful vocabulary in lieu of random tokens for prompt and prefix tuning, thereby boosting LLM performance on mathematical reasoning tasks.
\citet{srivastava2025debate} propose DTE, a ground truth-free training framework using multi-agent debates and a Reflect-Critique-Refine strategy to enhance LLM reasoning, achieving notable accuracy gains and strong cross-domain generalization.
Further, \citet{alazraki-rei-2025-meta} introduce a meta‑reasoning‑based tool selection framework, a two‑stage system first performs meta‑reasoning over the given task and then leverages a custom, fine‑tuned language‑modeling head to generate candidate tools, thereby substantially improving mathematical reasoning performance.

\begin{takeaways}
	\ \paragraph{Takeaways} \textit{(1) LLMs offer a unified generative solution paradigm for various NLP tasks.
	(2) LLMs in NLP tasks still have a certain gap from smaller supervised learning models.
	(3) Continuing to fine-tune LLMs on NLP tasks  bring substantial improvements.}
\end{takeaways}

\section{Future Work and New Frontier}
In this section, as shown in Figure~\ref{fig:future}, we highlight some new frontiers, aiming to inspire further innovations and groundbreaking advancements in the near future.

\subsection{Multilingual LLMs for NLP}
Despite the significant success of LLMs in English NLP tasks, there are over 7,000 languages worldwide. How to extend the success of English-centric LLMs to NLP tasks in other languages is an important research question~\citep{qin2024multilingual,winata-etal-2023-decades,li2025language,wang2025x,zhang2025cchall}.
Inspired by this,
Researchers have made efforts to enhance the multilingual LLM through parameter-tuning strategies, including multilingual pretraining~\cite{xue2021mt5,workshop2022bloom,chen2023monolingual,cahyawijaya2023instruct}, supervised fine-tuning~\citep{chen2023monolingual,cahyawijaya2023instruct,li2023eliciting}, and reinforcement learning~\citep{bajpai2025multilingual}. Other studies focus on cross-lingual alignment via prompting, using few-shot approaches~\citep{winata2021language,shi2022language,lin-etal-2022-shot,tanwar-etal-2023-multilingual} and zero-shot instructions~\citep{qin2023cross,huang-etal-2023-languages} to enhance alignment.

Two main challenges in this direction are as follows: (1) \textbf{\textit{Enhancing Low-Resource Language Performance}:}
Due to poor performance in low-resource languages, 
how to build universal multilingual LLMs that achieve promising performance in NLP tasks across languages is a direction worth exploring. (2) \textbf{\textit{Improving Cross-lingual Alignment}:} The key to multilingual LLMs is improving the alignment between English and other languages. Effectively achieving this alignment is critical for ensuring optimal performance in cross-lingual NLP tasks, making it a challenging yet essential area for advancement.

\subsection{Multi-modal LLMs for NLP}
The current LLMs achieve excellent performance in text modality. However, integrating modalities is one of the key ways to achieve AGI~\citep{huang2019multimodal,wang2025multimodal,li2025survey,peng2025skywork}. Therefore, a lot of work has begun to explore multi-modal LLMs for multi-modal NLP tasks~\citep{lu2022learn,qin2023mmsd2,qin2023cliptext,yang2023dawn,fei2024video,qin2024factors,zhang2025vitcot}.

The primary challenges in this field are:
(1) \textit{\textbf{Complex Multi-modal Reasoning:}} Currently, most multi-modal LLMs focus on simple multi-modal reasoning, like recognition~\citep{wang2023cogvlm,liu2023improved}, while neglecting complex multi-modal reansoning~\citep{chen2024m3cot,yang2023mm,lu2023mathvista}. Therefore, how to effectively explore complex multi-modal reasoning for NLP is a crucial topic~\citep{zhang2023multimodal,cheng2025comt,wang2025multimodal,cheng2025visual}. (2) \textit{\textbf{Effective Multi-modal Interaction:}} Existing methods often simply focus on adding direct multi-modal projection or prompting  to LLM for bridge multi-modality gap~\citep{wang2023cogvlm,liu2023improved,wu2023role,mitra2023compositional,wang2024s3}. Crafting a more effective multi-modal interaction mechanism in the inference process of multi-modal LLMs to solve NLP tasks is an essential problem.

\subsection{Tool-usage in LLMs for NLP}
While LLMs have shown success in NLP tasks, they can still face challenges when applied in real-world scenarios~\citep{qin2023toolllm,hu2023tree}.
Therefore, a lot of work focuses on exploring utilizing LLMs as central controllers to enable the usage or construction of tools and agents to solve practical NLP tasks~\citep{shinn2023reflexion,wang2023survey,zhu2023ghost,hu2024hiagent, zhang2025multi, yue2025masrouter}. 

The primary concerns are:
(1) \textit{\textbf{Appropriate Tool Usage:}} Current works always consider static tool usage, neglecting to choose appropriate tools to use. Identifying the correct tools and using them accurately is a key issue in solving NLP tasks efficiently.
(2) \textit{\textbf{Efficient Tool Planning:}} Current works still focus on the usage of a single tool for NLP tasks. 
Motivated by this, there is a pressing need for NLP tasks to achieve an efficient tool chain that leverages multiple tools in a coordinated manner. For example, when facing Task-oriented Dialogue tasks, we can use three tools: booking flight tickets, booking train tickets, and booking bus tickets. Then, how to collaborate to make the trip time as short as possible and the cost as low as possible is a typical problem in effective tool planning.

\subsection{X-of-thought in LLMs for NLP}
When LLMs solve complex NLP problems, they often cannot directly give correct answers and require complex thinking. Therefore, some works adapt X-of-thought (XoT) for advanced logical reasoning. XoT primarily focuses on refining the model’s ability to process and reason through complex logic, ultimately aiming to improve the overall performance and accuracy in solving challenging NLP tasks~\citep{kojima2022large,zhang2022automatic,qin2023cross,yao2023tree,chen2022program,lei2023boosting,zhang2024wrong}.

Key challenges in this direction include:
(1) \textit{\textbf{Universal Step Decomposition:}} How to develop a method for universally applicable step decomposition to generalize LLMs to various NLP tasks is the core challenge of XoT.
(2) \textit{\textbf{Prompting Knowledge Integration:}} 
Diverse promptings enhance model performance across various scenarios. How to better integrate the knowledge of different XoT to solve NLP problems is an important direction.

\subsection{Hallucination in LLMs for NLP}
During solving the NLP tasks, LLMs inevitably suffer from the hallucinations where LLMs produce outputs that deviate from world knowledge~\citep{muhlgay2023generating,min2023factscore}, user request~\citep{adlakha2023evaluating}, or self-generated context~\citep{liu2022token}. This deviation harms the reliability of LLMs in practical scenarios.

The primary challenges in hallucination are:
(1) \textit{\textbf{Efficient Hallucination Evaluation}}: How to find appropriate and unified evaluation benchmarks and metrics for LLMs in various NLP tasks is a key challenge.
(2) \textit{\textbf{Leveraging Hallucinations for Creativity}}: Hallucinations can often stimulate certain creative abilities. How to leverage hallucination to stimulate creativity and generate better innovative knowledge is an interesting topic.

\subsection{Safety in LLMs for NLP}
Applying large models to downstream NLP tasks also raises inevitable safety concerns,
including copyright issues~\citep{chang2023speak}, hate toxicity~\citep{hartvigsen-etal-2022-toxigen},  social bias~\citep{Wan2023BiasAskerMT,Dhamala2021BOLDDA} and psychological safety~\citep{Huang2023EmotionallyNO}. Inspired by this, a growing body of research has emerged, focusing on ensuring the safety of LLMs for various NLP tasks~\citep{ganguli2022red,sun2023safety,pan2025hidden,yu2025survey}.

The main challenges to safety in LLMs are:
(1) \textit{\textbf{Safety Benchmark Construction:}} 
Currently, there are few security-related benchmarks for LLM on various NLP tasks.
Establishing effective safety benchmarks is a critical objective in this area.
(2) \textit{\textbf{Multilingual Safety Risks:}} 
LLM suffers more safety risks across languages and cultures~\citep{xu2025survey}. 
Identifying and mitigating these risks in a multilingual context is a significant challenge.

\subsection{Long Chain-of-Thought in LLMs for NLP}
Long Chain-of-Thought (Long-CoT) extends standard CoT prompting by allowing models to reason more deeply, explore multiple solution paths, and reflect on intermediate outcomes instead of following a single linear chain of thought~\citep{chen2025towards,li2025system,jaech2024openai}. By organizing reasoning into hierarchical levels or segmented sub-chains, Long-CoT equips large language models to address complex NLP challenges and compositional reasoning tasks beyond the reach of conventional CoT~\citep{li2023symbolic,peifeng2023scott,chen2025ecm,wei2022chain,lyu2023faithful,yao2023tree,zeng2025FSDrive}. Recent innovations integrate reflective mechanisms~\citep{shinn2023reflexion,renze2024self}, inference-time scaling techniques~\citep{wang2023selfconsistency,balachandran2025inference,wu2024inference}, and reinforcement-learning enhancements~\citep{guo2025deepseek,yu2025dapo,yuan2025vapo,seed2025seed,duan2025efficient}.

Key challenges in this direction include:
(1) \textit{\textbf{Adaptive Reasoning Length Control:}}
Selecting the appropriate depth and breadth for each sub‑chain is challenging~\citep{sui2025stop,feng2025efficient,hou2025thinkprune}. If a sub‑chain is too shallow, the model may overlook critical intermediate abstractions; if it is too deep, it risks propagating errors or exceeding token limits~\citep{chen2024unlocking,chen2025rbf++}.
(2) \textit{\textbf{Interactive Reasoning:}}
Enabling a dynamic, iterative problem‑solving process, where models pose clarifying questions~\citep{qi2023art}, integrate external feedback~\citep{hu2023tree,paul2024refiner}, and refine intermediate steps~\citep{madaan2023self}, remains insufficiently explored~\citep{li2025beyond,chen2025towards}. Such interactive chains could substantially improve performance and accuracy in tasks requiring real‑time adaptation~\citep{yao2023react,chen2024essential}.
\section{Conclusion}
In this work, we make the first attempt to offer a systemic overview of LLMs in NLP, introducing a unified taxonomy of parameter-frozen paradigm and parameter-tuning paradigm. Besides, we highlight new research frontiers and challenges, hoping to facilitate future research. Additionally, we maintain a publicly available resource website to track the latest developments in the literature.
We hope this work can provide valuable insights and resources to build effective LLMs in NLP.

\section*{Acknowledgments}
This work was supported by the National Natural Science Foundation of China (NSFC) via grant 62306342, 62236004, 62206078 and 62476073. This work was supported by the Scientific Research Fund of Hunan Provincial Education Department (24B0001). This work was sponsored by the Excellent Young Scientists Fund in Hunan Province (2024JJ4070), the Science and Technology Innovation Program of Hunan Province under Grant 2024RC3024 and CCF-Zhipu Large Model Innovation Fund (NO.CCF-Zhipu202406).
This work was carried out in part using computing resources at the High Performance Computing Center of Central South University.

\bibliography{ref}

\begin{biography}{FCS-250472.R2-fig5}
Libo Qin received his Ph.D. degree in computer science from the Harbin Institute of Technology, China. He is a Professor at Central South University. His research interests include natural language processing and large language models.
\end{biography}

\begin{biography}{FCS-250472.R2-fig6}
Qiguang Chen is a PhD student in Harbin Institute of Technology (HIT), China. His research fields include natural language processing and complex reasoning.
\end{biography}

\begin{biography}{FCS-250472.R2-fig7}
Xiachong Feng is a Postdoctoral Researcher at the University of Hong Kong, holding a Ph.D. from the Social Computing and Interactive Robotics Research Center at Harbin Institute of Technology. He was also a visiting student at National University of Singapore. His research focuses on large language models (LLMs) and social agents, with publications in top-tier venues like ACL, TASLP, and TMLR. Awarded the National Scholarship three times, he has also received the CCL 2021 Best English Paper Award, TMLR Survey Award, and ICASSP 2023 MUG Challenge championship. He actively contributes to the academic community as a PC member/Area Chair for ICML, ICLR, and ACL Rolling Review. His work bridges AI, NLP, and human-agent interaction.
\end{biography}

\begin{biography}{FCS-250472.R2-fig8}
Yang Wu is a Ph.D. graduate in Computer Science from Harbin Institute of Technology (2024), with research expertise in improving the planning and cross-task generalization abilities of large language models for complex tasks. His doctoral work received the Best Paper Award at IEEE Multimedia 2021.
\end{biography}

\begin{biography}{FCS-250472.R2-fig9}
Yongheng Zhang is a master student at Central South University. His primary research interests include large language models and multimodal reasoning.
\end{biography}

\begin{biography}{FCS-250472.R2-fig10}
Yinghui Li received the BEng degree from the Department of Computer Science and Technology, Tsinghua University, in 2020. He is currently working toward the PhD degree with the Tsinghua Shenzhen International Graduate School, Tsinghua University. His research interests include natural language processing and deep learning.
\end{biography}

\begin{biography}{FCS-250472.R2-fig11}
Min Li received her Ph.D. in Computer Science from Central South University, China, in 2008. She is currently a professor and the dean of the School of Computer Science and Engineering at Central South University. Her research focuses on computational biology, systems biology, and bioinformatics. She has authored over 100 technical papers in leading journals and conference proceedings, including Nature Communications, Genome Research, Genome Biology, Nucleic Acids Research, and Bioinformatics.
\end{biography}

\begin{biography}{FCS-250472.R2-fig12}
Wanxiang Che received his Ph.D. degree in computer science from the Harbin Institute of Technology (HIT), China, in 2008. He is a Full Professor in the School of Computer Science and Technology, HIT. His current research interests include natural language processing and large language models.
\end{biography}

\begin{biography}{FCS-250472.R2-fig13}
Philip S. Yu (Life Fellow, IEEE) is currently a distinguished professor and the Wexler chair of information technology with the Department of Computer Science, University of Illinois Chicago (UIC), Chicago, Illinois. Before joining UIC, he was with IBM Watson Research Center, where he built a world-renowned data mining and database department. He has authored or coauthored more than 780 papers in refereed journals and conferences. He holds or has applied for more than 250 U.S. Patents. His research interest include Big Data, including data mining, data stream, database, and privacy. He is a fellow of ACM. He was the editor-in-chief of the ACM Transactions on Knowledge Discovery from Data during 2011–2017 and IEEE Transactions on Knowledge and Data Engineering during 2001–2004. He was the recipient of several IBM honors including the two IBM Outstanding Innovation Awards, Outstanding Technical Achievement Award, two Research Division Awards, and 94th Plateau of Invention Achievement Awards.
\end{biography}

\end{document}